\documentclass[conference]{IEEEtran}
\usepackage{cite}
\usepackage{graphicx}
\usepackage{color}
\usepackage[bookmarks=false]{hyperref}

\usepackage{tikz}
\usepackage{lipsum}

\newcommand\copyrighttext{%
	\footnotesize \copyright2018 Personal use of this material is permitted. Permission from IEEE must be obtained for all other uses, in any current or future media, including reprinting/republishing this material for advertising or promotional purposes, creating new collective works, for resale or redistribution to servers or lists, or reuse of any copyrighted component of this work in other works.}
\newcommand\copyrightnotice{%
	\begin{tikzpicture}[remember picture,overlay]
	\node[anchor=south,yshift=10pt] at (current page.south) {\fbox{\parbox{\dimexpr\textwidth-\fboxsep-\fboxrule\relax}{\copyrighttext}}};
	\end{tikzpicture}%
}

\newcommand\blfootnote[1]{%
	\begingroup
	\renewcommand\thefootnote{}\footnotetext{#1}%
	\addtocounter{footnote}{-1}%
	\endgroup
}
\graphicspath{ {images/} }

\usepackage{tikz}
\usetikzlibrary{arrows}
\usepackage{xspace}
\usepackage{latexsym}
\usepackage{amsfonts}
\usepackage{amsmath}
\usepackage{amssymb}
\usepackage{color}
\usepackage{colortbl}
\usepackage{epsfig}
\usepackage{graphicx}
\usepackage{pbox}
\usepackage{paralist}
\usepackage{enumerate}
\usepackage[color,matrix,arrow,all]{xy}
\usepackage{comment}
\usepackage{booktabs}
\usepackage{balance}
\usepackage{stmaryrd}
\usepackage{pifont}
\usepackage{hhline}
\usepackage{listings}
\usepackage{array}
\usepackage{float}
\usepackage[flushleft]{threeparttable}
\usepackage{graphicx,wrapfig,lipsum}
\usepackage{subcaption}
\usetikzlibrary{arrows}
\usepackage{graphicx,dblfloatfix}

\DeclareMathAlphabet{\pazocal}{OMS}{zplm}{m}{n}

\usepackage{array}
\newcolumntype{M}{>{\begin{varwidth}{2cm}}l<{\end{varwidth}}}
\newcolumntype{L}[1]{>{\raggedright\let\newline\\\arraybackslash\hspace{0pt}}m{#1}}
\newcolumntype{C}[1]{>{\centering\let\newline\\\arraybackslash\hspace{0pt}}m{#1}}
\newcolumntype{R}[1]{>{\raggedleft\let\newline\\\arraybackslash\hspace{0pt}}m{#1}}

\newcommand{\ie}{{\em i.e.,}\xspace}
\newcommand{\eg}{{\em e.g.,}\xspace}

 \usepackage[framemethod=TikZ]{mdframed}
\usetikzlibrary{shadows}
\mdfdefinestyle{myframe}{%
    linecolor=black,
    outerlinewidth=0.6pt,
    roundcorner=10pt,
    innertopmargin=10pt,
    innerbottommargin=10pt,
    innerrightmargin=10pt,
    innerleftmargin=10pt,
    skipabove=0.6\baselineskip,
    }

\newcommand{\Ni}{({\em i})~}
\newcommand{\Nii}{({\em ii})~}

\newcommand{\Lb}{\pazocal{L}}

\def\ps@IEEEtitlepagestyle{%
	\def\@oddfoot{\mycopyrightnotice}%
	\def\@evenfoot{}%
}
\def\mycopyrightnotice{%
	{\footnotesize The copyright belongs to me!\hfill}
	\gdef\mycopyrightnotice{}
}

\begin{document}


\title{Two Birds with One Network: Unifying Failure Event Prediction and Time-to-failure Modeling}


\author
{\IEEEauthorblockN{Karan Aggarwal\IEEEauthorrefmark{1}}
\IEEEauthorblockA{University of Minnesota\\
Minneapolis, MN\\
\textit{aggar081@umn.edu }}
\and
\IEEEauthorblockN{Onur Atan\IEEEauthorrefmark{1}}
\IEEEauthorblockA{University of California\\
Los Angeles, CA\\
\textit{oatan@ucla.edu }}
\and
\IEEEauthorblockN{Ahmed K. Farahat, Chi Zhang, Kosta Ristovski, Chetan Gupta}
\IEEEauthorblockA{Industrial AI Laboratory, Hitachi America, Ltd. R\&D\\
Santa Clara, CA\\
\textit{firstname.lastname@hal.hitachi.com}}
}


\maketitle
\copyrightnotice


\begin{abstract}
One of the key challenges in predictive maintenance is to predict the impending downtime of an equipment with a reasonable prediction horizon so that countermeasures can be put in place. Classically, this problem has been posed in two different ways which are typically solved independently: (1) Remaining useful life (RUL) estimation as a long-term prediction task to estimate how much time is left in the useful life of the equipment and (2) Failure prediction (FP) as a short-term prediction task to assess the probability of a failure within a pre-specified time window. As these two tasks are related, performing them separately is sub-optimal and might results in inconsistent predictions for the same equipment. In order to alleviate these issues, we propose two methods: Deep Weibull model (DW-RNN) and multi-task learning (MTL-RNN). DW-RNN is able to learn the underlying failure dynamics by fitting Weibull distribution parameters using a deep neural network, learned with a survival likelihood, without training directly on each task. While DW-RNN makes an explicit assumption on the data distribution,  MTL-RNN exploits the implicit relationship between the long-term RUL and short-term FP tasks to learn the underlying distribution. Additionally, both our methods can leverage the non-failed equipment data for RUL estimation. We demonstrate that our methods consistently outperform baseline RUL methods that can be used for FP while producing consistent results for RUL and FP. We also show that our methods perform at par with baselines trained on the objectives optimized for either of the two tasks.
\end{abstract}

\blfootnote{\IEEEauthorrefmark{1}{The work was completed during K. Aggarwal and O. Atan's internships at Hitachi America, Ltd. R\&D}}

\section{Introduction}
\label{intro}
Predictive maintenance is a widely-adopted maintenance practice which is based on continually monitoring the condition of the equipment with the goal of determining the right maintenance actions to be taken at the right times. With the advance of Internet of Things (IoT) and its applications to industrial environments, data analytics algorithms can be applied to the data coming from equipment in real time in order to provide actionable insights about equipment health and performance and predict impending failures and downtime. The use of data-driven technologies for predictive maintenance increases equipment availability, reduces the cost of maintenance, and improve the safety of equipment operators.

One of the key problems in predictive maintenance is the prediction of equipment failures early enough so that the proper maintenance can be scheduled before the failure happens. This problem is posed in two ways: (i) Remaining Useful Life (RUL) estimation which estimates how much time is left in the useful life of the equipment, and (ii) Failure Prediction (FP) which estimates the probability that a failure is going to happen within a typically short time horizon~\cite{si2011remaining}. From a maintenance process perspective, RUL estimation is very useful for long-term planning of spare parts supply and maintenance scheduling. On the other hand, failure prediction is more useful for handling unexpected failures that might happen in a short time-span. In the literature, a number of techniques have been used for the RUL estimation and failure prediction, mostly utilizing the temporal models using time-series analysis~\cite{wu2007prognostics,pham2010estimation}, explicit degradation modeling ~\cite{si2012remaining},  hidden Markov models~\cite{dong2010tutorial}, and deep learning methods recently~\cite{gugulothu2017predicting,guo2017recurrent,zhao2017learning,zheng2017long,zhangequipment}. 

\begin{figure}[b!]
\vspace{-1.5em}
\centering
\includegraphics[clip, trim=0.5cm 3cm 0.5cm 3cm,width=0.42\textwidth]{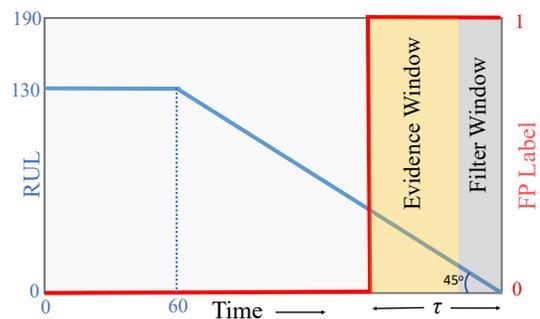}
\caption{Relationship between RUL and failure prediction (FP) labels on C-MAPSS with RUL capped at 130~\cite{zheng2017long}.}
\vspace{-1.5em}
\label{fig:rulfp}
\end{figure}

In the existing literature, failure prediction and RUL estimation are typically solved using separate models with different objective functions. In this work, we argue that using separate objective functions is sub-optimal and might result in inconsistent predictions. This argument is supported by multiple observations. \textbf{First}, the RUL estimation algorithms have a misplaced focus in trying to correct the estimation errors they make when the device is healthy ~\cite{zheng2017long,guo2017recurrent}.
Errors during this period of time should be assigned less weight compared to the estimation errors made close to the failure. For instance, a prediction that is off by ten days near the failure time is much worse than a prediction that is off by forty days a few months before failure. The focus of RUL methods is typically the latter case since it encounters much larger error. \textbf{Second}, if these models are applied in practice on the same 
component simultaneously, this might lead to inconsistent predictions, posing a dilemma for the decision maker.
 For example, the failure prediction model can predict that failure will happen in 5 days with 80\% probability, while the RUL model might predict that there are 10 days until the failure. 
\textbf{Third}, RUL estimation and failure prediction are related tasks --- \emph{RUL estimation is a long-term task while failure prediction is a short-term task} as shown in Fig~\ref{fig:rulfp}. By unifying the two, we can exploit their relationship and alleviate the previously mentioned issues. 

To address these challenges, we propose a unified formulation for the failure prediction and RUL estimation problems. This formulation will allow three modes of use: (1) failure prediction only; (2) RUL estimation only; (3) failure prediction and RUL estimation within a single model to ensure consistent predictions. We propose two models in this work: Deep Weibull model (DW-RNN) and Multi-task learning (MTL-RNN). Both models are using recurrent neural networks (RNN). The Deep Weibull model fits a Weibull distribution over the time-to-failure random variable using a survival analysis likelihood, with Weibull parameters being estimated by the network with appropriate transformations and training procedure. While the DW-RNN approach explicitly encodes the failure dynamics, we also propose an alternative approach of multi-task learning~\cite{caruana1997multitask} to implicitly learn such dynamics, so that 
we study our two tasks in one network. We present a novel constraint 
loss for leveraging the non-failed device data for RUL estimation in the multi-task framework.

We demonstrate that our methods outperform comparable baselines on the failure prediction task while being competitive with the baselines on the RUL estimation task. Typically, failure prediction is modeled as a classification task and RUL estimation as a regression task. One na\"\i ve approach to unify the two tasks is to use the estimated RUL for the failure prediction task by applying a threshold on the estimated RUL value. We demonstrate that this approach gives a sub-optimal performance for the failure prediction task. Failure prediction utilizes both the failure and non-failure data while RUL methods have only been used on failure data. Our proposed methods are capable of utilizing non-failure (censored) data without any need for a pre-prediction step of a dedicated unsupervised learning procedure~\cite{yoon2017semi}. This increases the applicability of the proposed algorithms as typically a large amount of data are available from non-failed equipment. In summary, this work makes the following contributions:
\begin{itemize}
\item To the best of our knowledge, this is the first attempt to unify the formulation of RUL estimation and failure prediction. Our method exploits the relationship between the two tasks and provide consistent predictions.
\item Ability to use the \textit{non-failure (censored) data} in an end-to-end approach which is not possible with most of the existing RUL estimation methods; and
\item Our methods achieves better performance on the harder task of failure prediction, by utilizing the RUL labels in comparison to the baselines that can be used to simultaneously achieve the two tasks.
\end{itemize}


\section{Related Works}
\label{sec:related}
In this section, we position the novelty of our approach in the predictive maintenance literature.   
RUL prediction and failure prediction are classical problems in predictive maintenance.
Most of the prior work in the literature has focused on the RUL prediction since the failure prediction can be inferred from the RUL estimates. RUL prediction problem has been studied using a 
variety of techniques from the machine learning literature mostly using sequential algorithms since RUL is a sequential task. 

Temporal models like auto-regressive 
models~\cite{wu2007prognostics,pham2010estimation}, diffusion processes~\cite{si2012remaining},  hidden markov 
models~\cite{dong2010tutorial}, and sequential deep learning methods 
recently~\cite{gugulothu2017predicting,guo2017recurrent,zhao2017learning,zheng2017long}. Other methods utilize 
frequency domain analysis on the sensor data~\cite{qiu2006wavelet,wang2008fault}. Recently, \emph{Deep learning techniques}~\cite{zhao2019deep} based on Long-Short Term Memory (LSTM) architecture have shown to produce 
state-of-the-art results~\cite{zheng2017long,guo2017recurrent,filonov2016multivariate,gugulothu2017predicting} owing to their superior feature 
extraction from the sequential data without any need for the  hand-crafted features.
For the failure prediction task, many methods have been proposed based on different off-the-shelf classification methods like SVM, Logistic Regression or Random 
Forests~\cite{aussel2017predictive,botezatu2016predicting}. Owing to the similarity of the problem to the \textbf{survival analysis} commonly used in the health-care setting~\cite{wang2017machine} to model time-to-death events, survival analysis techniques have been used. Most of these techniques use Weibull distribution~\cite{jing2016prediction} 
assumption on the \emph{time-to-failure} event since Weibull models a linear hazard rate, which corresponds to the linear degradation assumption that most of the literature makes. Other survival analysis techniques that have been used involve proportional hazards model~\cite{kumar1994proportional}, which are not relevant to the RUL estimation which involves
exact prediction of the time-to-failure event and not a relative risks framework. 

To the best of our knowledge, \emph{this is the first work that unifies the two prediction tasks of failure prediction and remaining useful life}. In this work, we present two methods: Deep Weibull network and a Multi-
task learning framework. The Deep Weibull network learns the Weibull distribution parameters as a function of deep 
LSTM network conditioned on the input sensor signal. The closest work to our Deep Weibull method is by Martinsson~\cite{martinsson2016wtte}. We introduce novel transformations to the distribution parameters learned
from the network as described in next section, and a pre-initialization procedure of the network which is absent in Martinsson's work, making it extremely hard to train and of limited practical use. 
Multi-task learning~\cite{caruana1997multitask,collobert2008unified} has been used in machine learning for learning multiple tasks together. The novelty of our multi-task learning architecture lies with an ability to learn from the non-failure (censored) data through a constraint loss where the labels for the RUL estimation task are not available. Additionally, both our proposed methods can learn from the non-failure data in an end-to-end 
learning procedure without any need 
for the pre-training procedures proposed recently for the RUL 
prediction~\cite{hu2017systematic,yoon2017semi,malhotra2016multi}.

\section{Problem Formulation}
\label{sec:formulation}
\begin{table}[b!]
	\vspace{-2em}
\caption{Notation Table}
	\vspace{-1.5em}
\begin{center}
 \scalebox{1}{
\begin{tabular}{ |c|p{7cm}| }
\hline
 \textbf{Notation} & \textbf{Explanation}  \\ 
 \hline
  $\mathcal{D}$ & Data with the sensor and failure or censoring information \\\hline
  $T$ & Time of failure random variable \\\hline
  $S(t)$ & Survival Function  \\\hline
  $f(t)$ & Probability density function\\\hline
  $F(t)$ & Cumulative density function \\\hline
  $h(t)$ & Hazard rate function  \\\hline
  $c_p$ & Time until which the sensor data is observed (censoring time) for device p \\\hline
  $t_g$ & Ground truth value of time-to-failure (RUL) for failed devices. Equals to time-to-censoring $c_p-t$ for the non-failed device $p$ at time t\\\hline
  $\tau$ & Failure prediction horizon\\\hline
  $\mathbb{I(\cdot)}$ & Indicator function \\\hline
  $N$ & Number of devices\\\hline
  $\mathbf{x}_{p}$ & Input time-series data $[\mathbf{x}_1,\mathbf{x}_2,\dots,\mathbf{x}_{c_p}]$ for device $p$ \\\hline
  $t_{p,f}$ & Failure time for a device $p$\\\hline
  $\delta_{p,t}$ & Label indicating whether the device $p$ suffers a failure until time $t$ \\\hline
  $\lambda$ & Scale parameter for the Weibull distribution \\\hline
  $k$ & Shape parameter for the Weibull distribution \\\hline
  $f^{p,t}$ & Ground truth failure prediction label, equals one if there is a failure in time $[t,t+\tau]$ \\\hline
$\mathrm{\widehat{RUL}}_{DW}$ & RUL value prdicted by the Deep Weibull model\\\hline
$\mathrm{\widehat{RUL}}_{MT}$ & RUL value prdicted by the Multi Task Learning model\\\hline
\end{tabular}
}
\end{center}
\vspace{-0.5em}
\end{table}

\begin{figure*}[t!]
\centering
\includegraphics[clip, trim=0.4cm 4.5cm 1.1cm  5cm,width=0.8\textwidth]{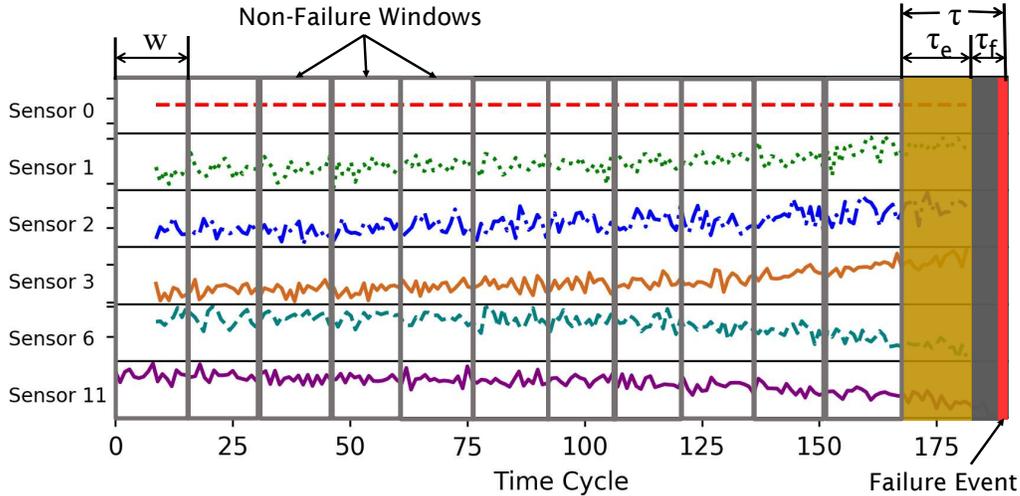}
\caption{An example from C-MAPSS, a turbofan engine degradation dataset from NASA~\cite{ramasso2014performance} showing sensor data and the proposed window scheme used in this work. We first divide our data into windows (of size $\mathsf{w}$).  The \textit{filter window} $\tau_f$  is removed from the data and windows falling in the \textit{evidence window} $\tau_e$ are labeled as the failure windows where the failure is obeserved within a horizon of $\tau$.  }
\vspace{-1.5em}
\label{fig:windows}
\end{figure*}

\noindent In this section, we formally define our problem. Given data $\mathcal{D} = \{\mathbf{x}_{p},c_p\}_{p=1}^{N}$, where $p$ is one of the $N$ devices, $\mathbf{x}_p$ are the
sensor observations for the device, and $c_p$ is the censoring random
variable for the device $p$. 
The censoring random variable defines the time until which the observations from the device 
were recorded. In case of failed devices, $c_p$ is same as the failure time $t_{p,f}$. For the devices that did not experience 
a failure until the end of observation time are referred to as \textit{censored} as per the standard survival analysis 
notation. The sensor observations are defined as the sequence of $d$-dimensional input:  $\mathbf{x}_p = 
[\mathbf{x}_1,\mathbf{x}_2,\dots,\mathbf{x}_{c_p}]$  with $\mathbf{x}_t \in \mathbb{R}^{d}$. The problem is to model 
the time-to-failure event, denoted henceforth by random variable $T$  with a probability density function given by $f(t|x)$. At time $t$, in the observation sequence we want 
to estimate the following two tasks:

 \textbf{Definition 1. Remaining Useful Life}: RUL is usually defined as time-to-failure, or mathematically speaking the expectation of the random variable $T$.
 	\vspace{-1em}
\begin{equation}
\label{eq:rul}
\mathrm{RUL} = \mathsf{\mathbb{E}}[T-t | T>t] = \int_{z=0}^{z=\infty} z f(z|x) dz
\end{equation}
 	
\textbf{Definition 2. Failure Prediction}: Failure prediction is defined as the probability of the device failing within a predefined number of day denoted by a horizon $\tau$. 
\begin{equation}
\label{eq:fp}
\mathrm{FP} = P(\tau\geq T-t|T>t) = \int_{z=0}^{z=\tau} f(z|x) dz
\end{equation}
\textit{While RUL is the expectation of the probability distribution, the failure prediction is the cumulative probability distribution within the horizon.}
Usually these problems have been modeled separately, \ie two different models for each of these problems. In this 
work, we make an attempt to jointly solve these problems in one model. Motivation behind creating a joint model being 
to produce consistent results for the two tasks while the task-specific models 
might produce mutually inconsistent results. We make the following key 
assumptions as a part of our formulation:
\begin{itemize}
\item The devices experience non-recoverable failures. While our methods can be easily extended to take into account the cases where devices can experience multiple failures with \eg a renewal process framework in survival analysis, it is not a subject of this study. 
\item The only form of censoring we observe is the right-censoring. There is no interval, \ie no missing observations. While we can use other key methods to handle the missing data, we do not model it explicitly.
\item Censoring or end-of-observation for the non-failed devices is non-informative and hence requires no special consideration over the time-to-failure.
\end{itemize}
\textbf{Failure Prediction Problem Setup:}
Next, we describe one of the key pieces of the problem --- setting the failure prediction task. While the RUL has ground truth labels available by assuming a linear relationship between degradation and end-of-life, for the failure prediction task only the presence or absence of failure is observed giving it a step function shape as shown in Fig~\ref{fig:rulfp}. We first divide each device's time-series into landmark windows of size $\mathsf{w}$, \ie  each window contains observations $x_t:x_{t+\mathsf{w}-1}$ as shown in Fig~\ref{fig:windows}. Assuming that the device $p$ fails at time $t_{p,f}$, we define the following terminologies:
\begin{itemize}
\item \textbf{Filter window}: Since prediction just before the failure time $t_f$ does not give enough warning in the realistic settings, we introduce the concept of filter window of size $\tau_f$.
\item \textbf{Evidence window}: Evidence window (size $\tau_e=\tau-\tau_f$) is the time during which the system undergoes (sudden) degradation and 
the failure probability is high. We assign the windows falling in the evidence window with failure labels since any window falling in the evidence window will experience a failure in the prediction horizon $\tau$. By formulating the failure prediction problem this way, helps us provide a timely warning which is critical in practice for scheduling maintenance and operational logistics. 
\end{itemize}
All the other windows are labeled as non-failure windows. Size of $\tau_e$ is domain dependent and a critical parameter in providing timely warning. For example, too small  $\tau_e$ does not provide much bandwidth for maintenance purpose depending on the domain while being highly predictive, while using too large value would make faulty assumptions on the degradation process with low predictive performance. In practice, 
for a domain like heavy machinery, small evidence window is not very useful, while for a domain like hard disk failures a relatively smaller value would suffice.

\emph{Note:} In the remaining paper, we drop the window size $\mathsf{w}$ for the ease of explanation. \textbf{We assume that the unit of time is the window size $\mathsf{w}$ unless specified otherwise.}

\section{Proposed Approach}
\label{sec:approach}
\noindent In this section, we present our methodology. First, we present preliminaries on survival analysis and show our proposed parametric Deep Weibull (DW) model. Following that we describe our non-parametric Multi-Task Learning (MTL) model.  

\subsection{Deep Weibull Model}
\noindent In this section, we first present the survival analysis framework and then present our Deep Weibull (DW-RNN) model.
\subsubsection{Survival Analysis Preliminaries}
\label{sec:surv}
In this section, we present the basic survival analysis framework. The time-to-failure is modeled with a random variable $T>0$. The survival function denotes the probability of survival of device until time $t$, or equivalently the probability of failure after time $t$:
\begin{equation}
S(t|x) = P(T>t|x)
\end{equation}

\noindent where $P(T>t|x)$ is the probability of failure time being greater than the current time $t$ conditioned on the input x. For the 
sake of simplicity, we drop the conditional to denote the probabilities throughout the paper. We can further write survival function $S(t)$ as:
 \begin{equation}
 S(t) = 1-F(t) = 1 - \int_{0}^{t} f(t) dt 
\end{equation}
 
\noindent $F(t)$ denotes the cumulative probability of failure until time $t$, and $f(t)$ is the probability density function of the failure event. The hazard rate $h(t)$ is defined as:
\begin{equation}
 h(t) = \lim_{dt\to 0} \frac{P(t\le T<t+dt | T \ge t)}{dt}  =  \frac{f(t)}{S(t)} 
\end{equation}

\noindent Usually, the choice of distribution is made based on the characteristics we want to model on the hazard rate function $h(t)$. $h(t)$ modulates the probability density 
function $f(t)$ based on the degradation or aging as characterized by the $S(t)$.
In the survival framework, the the remaining useful life 
and failure probability are same as given by Eq.~\ref{eq:rul} and ~\ref{eq:fp}. Let the failure flag $\delta_{p,t}$ denote whether the device $p$ had failed or not by time $t$, \ie
\begin{equation}
\delta_{p,t} = \mathbb{I} (t_{p,f}\le t)
\end{equation}
where the $t_{p,f}$ is the device's failure time. The likelihood of the observations $\mathcal{D}$ is given by:
\small
\begin{eqnarray}
P(\mathcal{D}) = \prod_{p=1}^{N} \prod_{t=1}^{c_p} \underbrace{p(T_p=t)^{\delta_{p,t}}}_\text{Failure at time $t$} \underbrace{S(T_p>t)^{1-\delta_{p,t}}}_\text{Not failed until time $t$} \\
 = \prod_{p=1}^{N} \prod_{t=1}^{c_p} p(T_p=t)^{\delta_{p,t}} (1-F(t))^{1-\delta_{p,t}}
\end{eqnarray}
\normalsize
The survival based negative log-likelihood is given by:
\small
\begin{eqnarray}
\label{ll}
-\mathrm{log}~P(\mathcal{D}) = -\sum_{p=1}^{N} \sum_{t=1}^{c_p} \Big \{ \delta_{p,t}\mathrm{log}~p(T_p=t)  \nonumber \\   
+ (1-\delta_{p,t}) \mathrm{log}~(1-F(t)) \Big \}
\end{eqnarray}
\normalsize
Next, we formulate this likelihood by parameterizing with a Weibull distribution. Please note that while there are other non-parametric 
survival analysis methods like Cox's regression they are not suitable for this task since they optimize the \emph{relative}
risks in the population and not the absolute time-to-failure prediction. We use Weibull to parametrize our likelihood since it is shown to be the distribution of choice in the predictive maintenance owing to its hazard rate $h(t)$ that is directly proportional to the ground RUL.  

\begin{figure}[t!]
\centering
\includegraphics[clip, trim=0.1cm 3cm 5.25cm 3cm,width=0.4\textwidth]{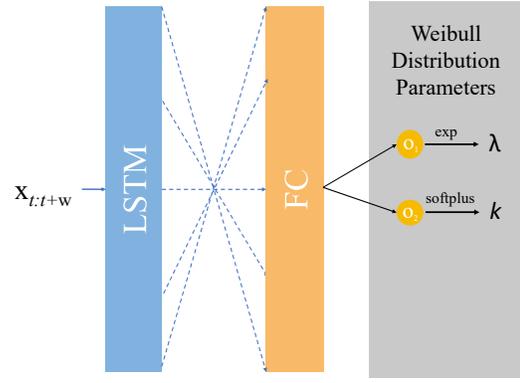}
\caption{Architecture for the Deep Weibull Network, where LSTM  and Full connected (FC) layers with exponential linear unit (ELU) activation can have multiple layers stacked. The output are Weibull distribution parameters $\lambda$ and $k$ .}
\vspace{-0.5em}
\label{fig:dw_arch}
\end{figure}

\subsubsection{Weibull Distribution}
The Weibull distribution is the most commonly used distribution in the prognostics literature for modeling time-to-failure. The distribution is characterized by two parameters, $\lambda$ (scale) and $k$ (shape). The probability density function and cumulative density function are given by:
\small
\begin{eqnarray}
p(t_{g},\lambda,k) =  \frac{k}{\lambda} \Big (\frac{t_{g}}{\lambda}\Big )^{k-1} e^{-\Big ( t_{g} /\lambda \Big )^{k}}\\
F(t_{g},\lambda,k) =  1 - e^{-\Big ( t_{g} /\lambda \Big )^{k}}\\
h(t_{g},\lambda,k) =   \frac{k}{\lambda} \Big (\frac{t_{g}}{\lambda}\Big )^{k-1}
\label{eq:whazard}
\end{eqnarray}
\normalsize
where, $t_{g}$ is the ground truth time-to-failure (RUL) for the devices that experienced failure. It equals time to censoring $c_p-t$ for the non-failed device $p$ at time t. While the survival analysis formulation in Section~\ref{sec:surv} is in terms of absolute time $t$, by using fundamental theory of expectation for change of variable, we reformulate in terms of $t_g$. For proof see~\cite{survproof}. The negative log likelihood objective from Eq~\ref{ll} can be written as:
\small
\begin{eqnarray}
-\mathrm{log}~P(\mathcal{D}) = -\sum_{p=1}^{N} \sum_{t=1}^{c_p} \Big \{ \delta_{p,t} \Big [ \mathrm{log}~\frac{k}{\lambda} + (k-1)\mathrm{log}~\frac{t_{g}^{p,t}}{\lambda} \Big ]  \nonumber \\ 
- \Big ( \frac{t_{g}^{p,t}}{\lambda} \Big )^{k}  \Big \} + \alpha \|\theta\|_{2}
\end{eqnarray}
\normalsize
 with $\alpha$ being the $l_2$ regularization for the network parameters $\theta$.
Owing to numerical stabilization needed during the training, we approximate the term $\Big ( \frac{t_{g}^{p,t}}{\lambda} \Big )^{k}$ with a polynomial expansion unto $4^{th}$ order:
\begin{eqnarray}
\Big ( \frac{t_{g}}{\lambda} \Big )^{k}  \approx  1 + \sum_{j=1}^{4} \frac{\Big [  k~\mathrm{log}~\frac{t_{g}}{\lambda} \Big ]^j}{j!}
\end{eqnarray}
\subsubsection{Deep neural network} We use a deep neural network based architecture as shown in Figure~\ref{fig:dw_arch} to estimate the likelihood above. For learning the distribution parameters, we model $\lambda$ and $k$ as a function of the input $x$ as:
\begin{eqnarray}
\lambda = q_{\theta}(x); k = r_{\theta}(x)
\end{eqnarray}

with $\theta$ being the neural network parameters, $q(\cdot)$ and $r(\cdot)$ are the function 
approximates of the deep neural network. Both the parameters are taken as the output from the last fully-connected layer of the neural network. However, owing to numerical stability issues since both the parameters work in different domains, we apply following transformations on the last layer outputs $\{o_1,o_2\}$:
\begin{eqnarray}
\lambda = \mathrm{exp}(o_1); k = \mathrm{softplus}(o_2)
\end{eqnarray}
where softplus activation~\cite{douglas1995recurrent} is the function $\mathrm{log}(1+\mathrm{exp}(o_2))$.
The network trains on the likelihood given above and learns the relationship between the input and distribution parameters as described above. 

\textbf{Inference:} During the inference phase, we estimate the parameters $\lambda$ and $k$ based on the input data from the 
trained network. From the estimated parameters $\widehat{\lambda}$ and $\widehat{k}$ , the RUL and failure prediction can be done with the following closed form expressions:
\small
\begin{equation}
\label{rul}
\mathrm{\widehat{RUL}}_{DW} = \mathsf{\mathbb{E}}[T-t | T>t] = \widehat{\lambda}~\Gamma \Big ( 1+ \frac{1}{\widehat{k}}\Big )
\end{equation}
\begin{equation}
\label{fp}
\mathrm{\widehat{FP}}_{DW} = P(\tau\geq T|T>t) = 1 - e^{-\Big ( \tau /\widehat{\lambda} \Big )^{\widehat{k}}}
\end{equation}
\normalsize
with $\Gamma(\cdot)$ being the extended factorial or gamma function.
This method makes an explicit assumption of a Weibull distribution over the time-to-failure, which might not be true in all the cases.
 Hence, we propose a second model based on multi-task learning with a motivation
for implicitly learning the distribution and exploiting the relationship between the two tasks, next.

\subsection{Multi-task Learning}
We propose another approach MTL-RNN that utilizes the multi-task learning~\cite{caruana1997multitask} framework that has been successfully used in a number of domains. 
The idea of multi-task learning is that if two or more tasks at hand are related to each other, a common useful feature space can be created by learning them jointly. As discussed in Section~\ref{sec:formulation},
 RUL is the expectation of the probability density of time-to-failure random variable over the time domain, while failure prediction is the cumulative distribution within a short horizon of the time-to-failure random variable. 
The graphical representation of the relationship is shown in Fig~\ref{fig:rulfp}. 

\begin{figure}[t!]
	\centering
	\includegraphics[clip, trim=0.1cm 3cm 1.25cm 3cm,width=0.4\textwidth]{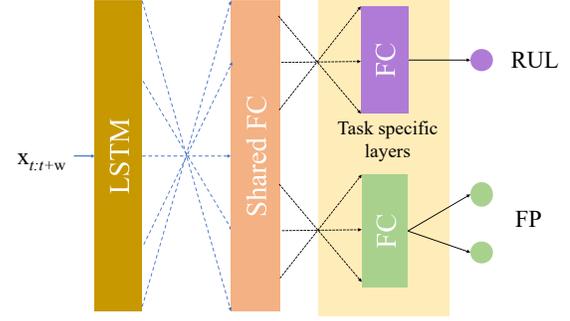}
	\caption{Architecture for the Multi-task Network, where LSTM  and Full connected (FC) layers  with exponential linear unit (ELU) activation can have multiple layers stacked. The output are FP and RUL values from task specific FC layers.}
	\vspace{-1.5em}
	\label{fig:mt_arch}
\end{figure}

 RUL is a longer term horizon task, while failure prediction is a shorter term horizon task. We hypothesize that learning them jointly helps the model learn the underlying distribution conditional on the input observations 
\emph{``implicitly"} without making an \emph{``explicit"} assumption like our Deep 
Weibull model.  We use a shared network for the two tasks, with task-specific layers emulating from the last shared 
layer.  The architecture schematic is shown in Figure~\ref{fig:mt_arch}. We describe the two loss functions used to train the two tasks:
\paragraph{Failure Prediction} The FP prediction task is trained with a cross-entropy loss:
\small
\begin{eqnarray}
\label{eq:class}
\Lb_c = \sum_{p=1}^{N} \sum_{t=1}^{c_p} \Big \{\mathbb{I}(f^{p,t} = 0) \log p(f^{p,t} = 0|\mathbf{x}, \theta) + \nonumber \\
+ \alpha_f \mathbb{I}(f^{p,t} = 1) \log p(f^{p,t} = 1|\mathbf{x}, \theta) \Big \}  
\end{eqnarray}
\normalsize
where, $f^{p,t}=1$ if device $p$ is in a failure state at time $t$, $\theta$ are the deep network's parameters, and $\alpha_f$ is the weighting parameter that can be tuned to give higher weighting to the failure class during training to account for the imbalance in the data.
$p(f^{p,t} = 0|\mathbf{x}, \theta)$ is the probability of non-failure state calculated using soft-max from the failure prediction layer. Let the input to the soft-max layer be $\gamma_\theta(x)$, where $\gamma_\theta(\cdot)$ is the neural network transformations of the input $x$. We can write the probability of failure to be:
\small
\begin{eqnarray}
\label{eq:softmax}
p(f = 1|\mathbf{x}, \theta) = \frac{\mathrm{exp}(W_1^{\mathsf{T}} \gamma_\theta(x))}{\mathrm{exp}(W_0^{\mathsf{T}} \gamma_\theta(x)) + \mathrm{exp}(W_1^{\mathsf{T}} \gamma_\theta(x))} 
\end{eqnarray}
\normalsize
with $W_0,W_1$ being the weights in the soft-max layer corresponding to be non-failure and failure class, respectively. 
\paragraph{Remaining Useful Life} The RUL prediction task for a device $p$ that failed can be formulated as a square error loss:
\small
\begin{equation}
\label{eq:reg}
\Lb_{r,p}^f =  \sum_{t=1}^{c_p} ( \widehat{RUL}_{MT}^{p,t} - t_{g}^{p,t})^2
\end{equation}
\normalsize
We denote the RUL estimation loss per device so as not to abuse the notation for the failed and non-failed devices. We treat the \emph{failed} and \emph{non-failed} devices separately since there is no ground truth RUL information about the non-failed devices. 

\textbf{Non-failed device (censored) data:} The multi-task formulation in Eq.~\ref{eq:mtl} assumes that we have the ground truth labels for both the tasks. However, that's rarely the case in the prognostics area. Usually, the number of non-failed devices is much higher compared to the failed devices. In practical situations, the numbers can be anywhere between $<$1\% to 5\% depending on the device's domain. In such a scenario we have no ground truth information about the RUL labels for the overwhelming majority of the data. The ability of survival analysis to take into account the censored data is what makes it an attractive option for prognostics. Most of the other works throw away the non-failed device data~\cite{zheng2017long}.
Yoon et al.~\cite{yoon2017semi,malhotra2016multi} used a pre-trained network with auto-encoder architectures to utilize the non-failed data for RUL prediction. 
In order to utilize the non-failure data, we want to make sure that the predicted RUL values for the non-failure 
data is greater than the censoring time $c_p$ of the device. The reason for this constraint is straight-forward: 
we know that the device didn't experience a failure until the end of observations at $c_p$, hence, it's failure 
time $t_{p,f}$ should be greater than $c_p$.  In order to enforce this constraint, we use the following 
objective for the non-failed devices:
\begin{equation}
\label{eq:creg}
\Lb_{r,p}^{nf} =  \mathbb{I} (t_{p,f}> c_p)~\sum_{t=1}^{c_p} ~\mathrm{max} (c_p - \widehat{RUL}_{MT}^{p,t},0)
\end{equation}
where, the $ \mathbb{I} (t_{p,f}\le c_p)$ is the indicator function set to 1 if the device $p$ fails within the 
observation period $c_p$ of the device. Hence, we can utilize the non-failure data to train our network on all 
instances for the failure prediction task, while ensuring that the RUL predicted is greater than the censoring 
time,  $c_p$. For the failed device data we learn on both the tasks with conventional objectives. Loss function 
for the RUL estimation task can be written as:
\small
\begin{eqnarray}
\vspace{-1em}
\label{eq:rulloss}
\Lb_{r} = \sum_{p=1}^{N}  \Big \{ \Lb_{r,p}^{f} + \Lb_{r,p}^{nf}  \Big \} \nonumber \\ 
=   \sum_{p=1}^{N}  \Big \{   \underbrace{\mathbb{I} (t_{p,f}\le c_p) \sum_{t=1}^{c_p} (\widehat{RUL}_{MT}^{p,t} - t_{g}^{p,t})^2}_\text{Device failure observed} \nonumber \\
 + \underbrace{\mathbb{I} (t_{p,f}> c_p)~\sum_{t=1}^{c_p}~\mathrm{max} (c_p - \widehat{RUL}_{MT}^{p,t},0)}_\text{Device failure not observed} \Big \}
\vspace{-1em}
\end{eqnarray}
\normalsize

\paragraph{Combined Loss} Using the loss functions in the two equations, Eq.~\ref{eq:reg},\ref{eq:class} we train the network jointly with a combined loss $\mathcal{J(\theta)}$:
\small
\begin{equation}
\label{eq:mtl}
\mathcal{J(\theta)} = \alpha_1 \Lb_c + \alpha_2 \Lb_{r} + \alpha_3 \|\theta\|_{2}
\end{equation}
\normalsize
with $\alpha_1$ and $\alpha_2$ being the hyper-parameters for adjusting the 
relative weights of the FP and RUL respectively, and $\alpha_3$ is the $l2$ regularization strength for the model parameters.



Hence, this learning procedure is \emph{fully supervised} for the failure prediction task, but \emph{semi-supervised} 
with a constraint described by Eq~\ref{eq:creg} for the RUL 
estimation owing to the shared layer structure of the multi-task learning framework. Hence, with the above formulation 
we not only try to ensure consistent predictions of RUL and failure prediction tasks but also leverage the non-failure data by using the 
shared layer structure.



\section{Experimental Settings}
\label{sec:settings}
In this section, we describe our datasets, evaluation metrics and the baselines used.
\subsection{Datasets}
We utilize two \textbf{public} datasets to evaluate our methods: C-MAPSS (Commercial Modular Aero-Propulsion
System Simulation) and Backblaze hard disk failure data. 

\paragraph{C-MAPSS} This is a turbofan engine degradation benchmark dataset\footnote{https://catalog.data.gov/dataset/c-mapss-aircraft-engine-simulator-data}~\cite{ramasso2014performance} provided by NASA that has been extensively used in the prognostics literature~\cite{zheng2017long,guo2017recurrent,jia2017assessment}. The dataset has four component datasets with each dataset having different operating conditions and failure modes. The data consists of three operational settings and sensor data collected from 21 sensors.
The test data contains the censored device data, however, with ground truth RUL values provided. 
\paragraph{Backblaze} We use the Backblaze hard disk\footnote{{https://www.backblaze.com/b2/hard-drive-test-data.html}} failure dataset for our task. Backblaze data-center maintains the record of the hard disk and any failures encountered by hard disk's manufacturer and make. Most of the prior works~\cite{botezatu2016predicting} have performed failure detection rather than making failure prediction as we formulated in 
Section~\ref{sec:formulation}. We used the data for the Seagate hard disk model \texttt{ST4000DM000} from Jan 2014 to June 2015, as it has the largest number of observations and the data collection methods were changed thereafter. The data consists of 26 S.M.A.R.T.  (Self-Monitoring, Analysis and Reporting Technology) 
features like temperature, error rates and other performance metrics aggregated for each day of observation. We perform test results only on the non-censored devices. 

\begin{table}[t!]
 \caption{Neural network architecture for our methods for the C-MAPSS and Backblaze datasets.}
\centering
 \scalebox{0.85}{\begin{tabular}{|c|c|} 
 \hline
 \multicolumn{2}{|c|}{C-MAPSS}\\\hline
Method & Architecture \\\hline
 FP-RNN & LSTM(128)-FC(32)-FC(16)-FC(2) \\
 RUL-RNN & LSTM(128)-FC(64)-FC(32)-FC(1)\\
 Deep Weibull &  LSTM(128)-FC(32)-FC(16)-FC(2)\\
 Multi-task RNN & LSTM(200)-FC(100)-FC(64)---[FP: FC(2),RUL: FC(32)-FC(1)] \\
 \hline\hline
  \multicolumn{2}{|c|}{Backblaze}\\\hline
Method & Architecture \\\hline
 FP-RNN & LSTM(64)-FC(16)-FC(2)\\
 RUL-RNN & LSTM(64)-FC(32)-FC(1)\\
 Deep Weibull &  LSTM(64)-FC(32)-FC(16)-FC(2)\\
 Multi-task RNN & LSTM(100)-FC(64)-FC(16)---[FP: FC(2),RUL: FC(16)-FC(1)] \\
 \hline

 \end{tabular}
 }
\vspace{-1em}

 \label{table:arch}
\end{table}

\subsection{Data Preprocessing}
The key assumption as widely used in the literature is that RUL is a linear function of time as interpolated back from the failure time $t_{p,f}$. In practical terms, though this assumption would be faulty since the device degradation is much more severe near the failure time. However, that is a subject of another problem commonly tackled with health indicator prediction. Recent works~\cite{heimes2008recurrent,babu2016deep,zheng2017long} have shown that using a piece-wise linear function of RUL with a capped value of maximum RUL and a linear degradation thereafter is a better modeling practice. In accordance with the practice for C-MAPSS we set the maximum RUL value to 130 operating cycles. For the Backblaze dataset, we use a value of 50 days.

\textbf{Window extraction:} We divide the device's sensor measurements data into fixed size landmark windows that are fed as an input to the deep network composed of LSTM units as described in Section~\ref{sec:formulation}. The window size $\mathsf{w}$ is a hyper-parameter we tuned during our training. The filter windows $\tau_f$ 
are fixed based on a reasonable assumption on the domain --- we fix filter window size to be 5 days for the  C-MAPSS and 4 days for the hard disk dataset. The evidence window size 
$\tau_e$ are selected empirically and with domain knowledge.
We set the evidence window size to be 20 for C-MAPSS and 12 for the hard disk data. 

\textbf{Class Balancing:} As mentioned previously, the datasets are highly skewed with very few devices experiencing a failure. In order to tackle this issue, we used random sampling methods to balance the number of failed devices to non-failed devices. We found that up-sampling and down-sampling performed similarly; hence, we show the experimental results with down-sampling owing to smaller computational time. However, one major problem that still stands is the fact that only a minuscule number of windows posses the failure label 
(evidence windows) --- 0.85\% for C-MAPSS and 4\% for Backblaze.  We use the class weighting as hyper-parameter for the multi-task learning approach to tackle this problem. 

\subsection{Evaluation Metrics}
We use the following evaluation metrics for evaluating the performance of our models:
\paragraph{RMSE} For RUL prediction we use RMSE (Root Mean Square Error) as has been widely used in the literature. It is computed as follows:
\begin{equation}
\label{eq:rmse}
\mathrm{RMSE} =  \sqrt{\frac{1}{n} \sum_{i=1}^{n} ( \widehat{RUL}^{i} - t_{g}^{i})^2}
\end{equation}
 with $n$ being the total number of windows in the data RUL prediction is being evaluated on.
\paragraph{AUC} For the failure prediction task, we use area under the receiver-operator curve (ROC).

However, the AUC-ROC can be very misleading for the highly skewed 
data~\cite{davis2006relationship} like ours, especially when the focus is on the prediction of the minority class. As suggested by Davis and 
Goadrich~\cite{davis2006relationship}, we also use the AUC-PR (Area under Precision Recall Curve).   
AUC-PR shows much more differential for skewed datasets as reflected in our experiments.

\begin{figure*}[b!]
	\centering
	\begin{subfigure}{.4\textwidth}
		\centering
		\includegraphics[width=.8\linewidth]{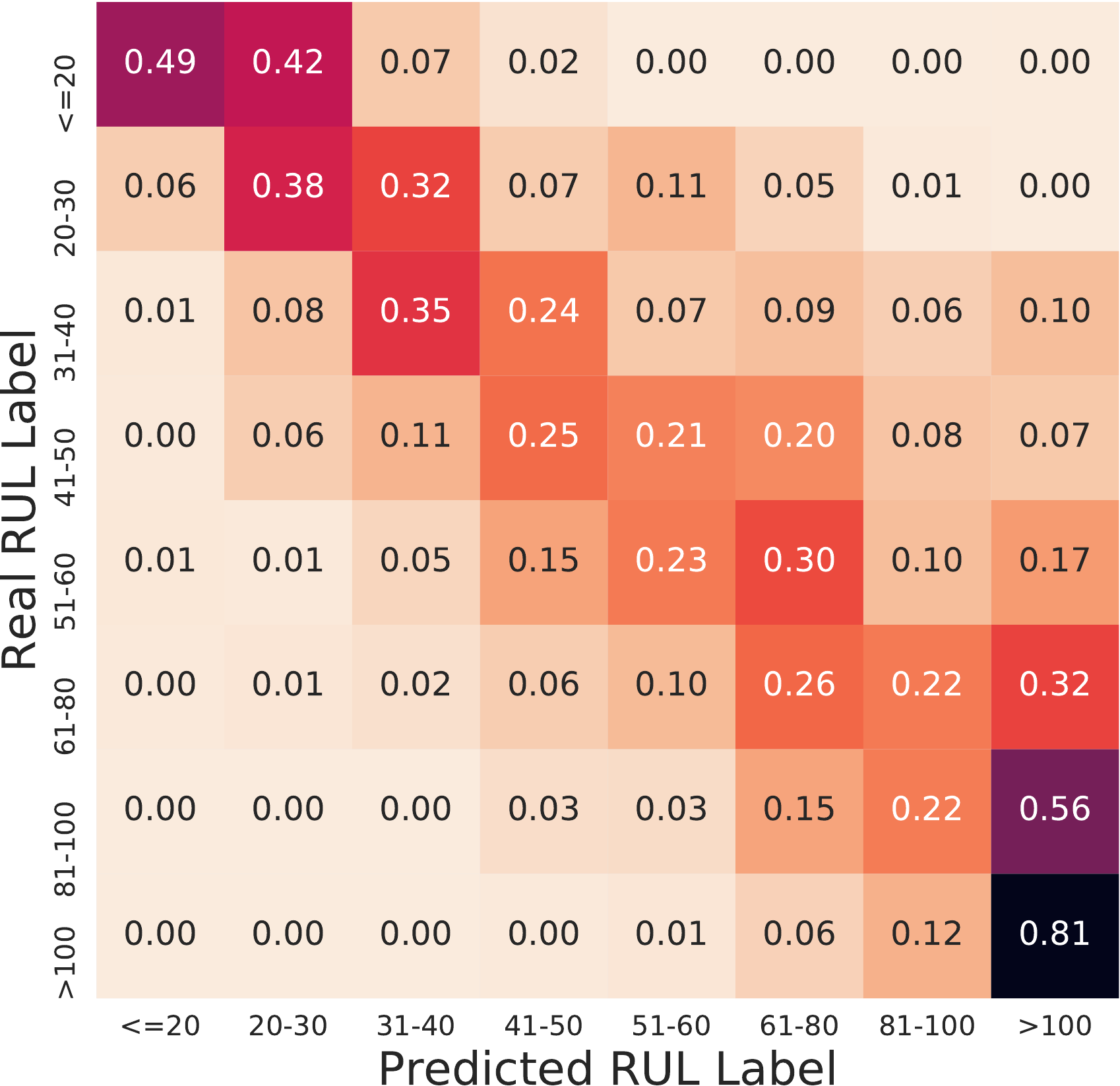}
		\label{fig:trans_crf}
	\end{subfigure}
	\begin{subfigure}{.4\textwidth}
		\centering
		\includegraphics[width=.8\linewidth]{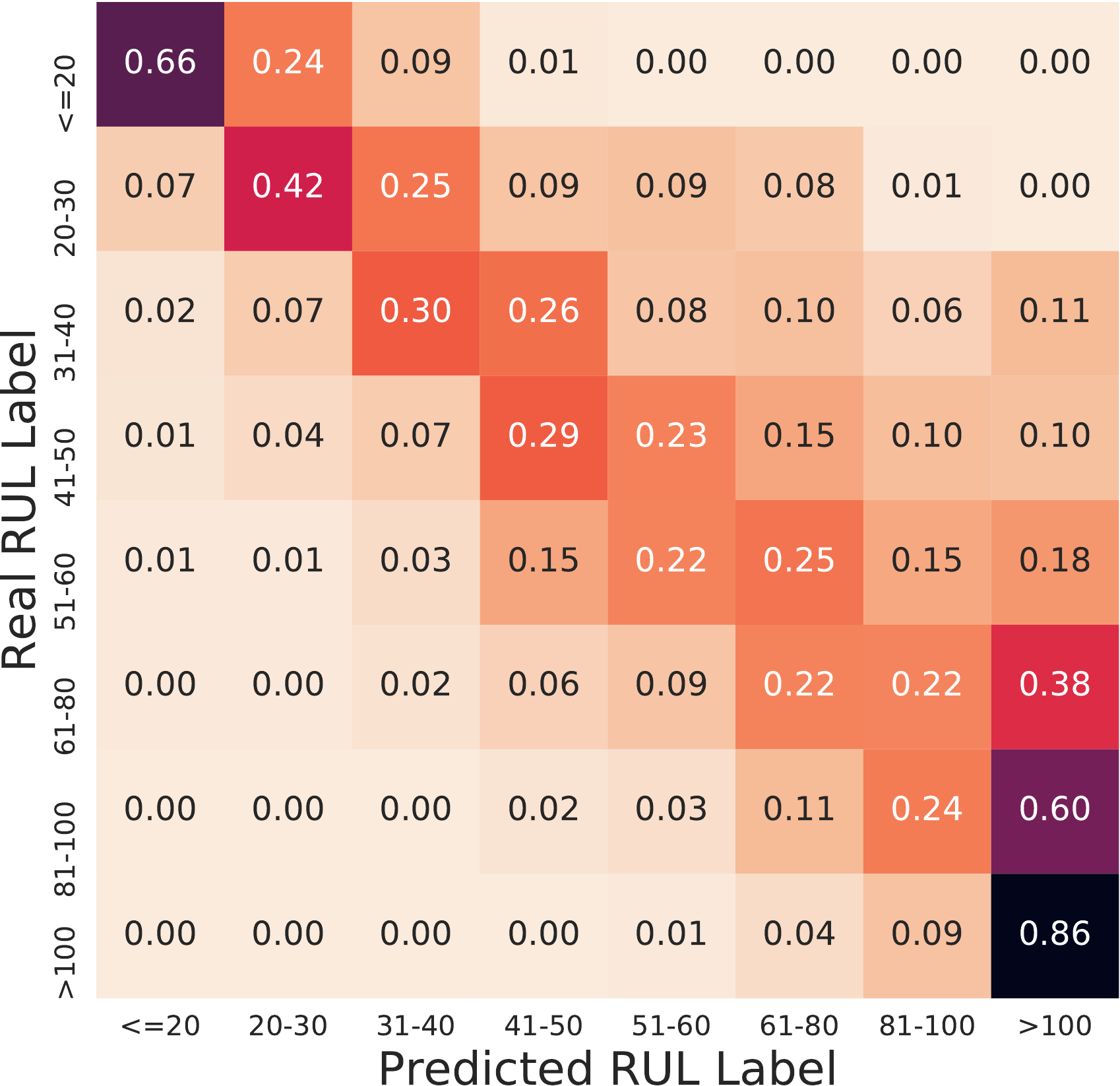}
		\label{fig:conf}
	\end{subfigure}%
	\caption{Confusion matrices by binning the real RUL values and predicted RUL values from the baseline RUL-RNN (left) and MTL-RNN (right) for the C-MAPSS FD01. }
	\label{fig:conf_matrix}
\end{figure*}

\subsection{Baselines Methods}
We use the following baseline methods:

\textbf{\Ni Failure Prediction baselines}:
\paragraph{FP-RNN} This is based on the same neural network architecture as our methods, but with just one output based on the soft-max layer as Eq~\ref{eq:class}.
\paragraph{Random Forest} We use the Random Forest classifier for the failure prediction task over each window.
\paragraph{Logistic Regression} LR for each window's label.
\paragraph{ Support vector machine} SVM with a  linear kernel for each window.

\textbf{\Nii RUL and Failure Prediction baselines}:
Since all the methods that can perform the RUL prediction can be used for the failure prediction by using the horizon time threshold over predicted RUL values, we use the following baselines: 
\paragraph{RUL-RNN} Similar to FP-RNN, this network is just trained on the RUL prediction task in accordance to the Eq~\ref{eq:reg}.
\paragraph{SVR}We use the support vector regression, with a linear kernel.

We report the prediction metrics for all our baselines and additional failure prediction metrics from the RUL prediction models. Next, we describe the optimization  
we had to use for training the Deep Weibull network.

\subsection{Pre-training for Deep Weibull Network}
Given that we are trying to fit a distribution over the neural network to learn the parameters conditioned on the input sensor data, we found it very challenging to stabilize the training. We had to employ pre-training of the deep network with the RUL estimation task by calculating the RUL with Eq.~\ref{rul} and using the loss described in ~\ref{eq:reg} to train the network. 
In other words, we first 
train the network with the RUL estimated through Eq~\ref{rul} using the loss from Eq~\ref{eq:reg} until convergence through early stopping on the RMSE metrics. Once the network learns the parameters on the RUL estimation task, we use the likelihood in Eq~\ref{ll} to train the network for Weibull distribution.

\subsection{Experiments and Hyper-parameter Tuning}
We use the test data provided in C-MAPSS for testing. For the BackBlaze dataset, we use 80\% of the data for training and 20\% for testing.  For both the datasets, we use randomly chosen 30\% of the training data as the validation set. All our models were implemented in TensorFlow~\cite{abadi2016tensorflow}, and the baselines were evaluated using scikit-learn~\cite{pedregosa2011scikit}. We use Adam optimizer with a learning rate of 1$e^{-3}$. We use early stopping criteria with optimization on AUC-PR for the joint and failure prediction tasks, and RMSE for the RUL regression tasks. This whole procedure is run 10 times, and we report the mean of metrics for these 10 runs. For the LSTM layer the number of hidden units was chosen 
from the set $\{64,100,128,200,256\}$, while for the fully connected layer size was chosen from $\{8,32,64,128,200\}$ with a dropout~\cite{srivastava2014dropout} probability of 0.1. For the C-MAPSS data we tune between the evidence window $\tau_e$ of $\{10,20,30\}$ and window size $\mathsf{w}$ of \{5,10\}. 
We finally chose evidence window of 20 and 10-cycle long windows.  While for the Backblaze data we
experiment between evidence windows of $\{7,8,10,12,14\}$ 
and window sizes of $\{3,4,5,8,10\}$, finally using an evidence window of 12 and window size of 4. For the class weighting factor, we use a 
factor of 1000 for the C-MAPSS and 500 for the Backblaze dataset after careful tuning. We found that using relative task weights of $\alpha_1=1$ (FP) and $\alpha_2=1$ (RUL) works well for both the datasets.

\begin{table}[t!]
\centering
 \caption{RMSE scores for the RUL prediction task using different methods.}
 \begin{tabular}{|c|c|c |c|c|c|} 
 \hline
&  \multicolumn{4}{|c|}{C-MAPSS} & \multicolumn{1}{c|}{BackBlaze} \\\hline
Method & \multicolumn{1}{|c|}{FD001} & \multicolumn{1}{c|}{FD002} & \multicolumn{1}{c|}{FD003} & \multicolumn{1}{c|}{FD004} &   \\ \hline
 SVR & 22.86  & 26.96 & 23.54  & 28.40 & 34.91\\
 RUL-RNN & \bf{21.29}& \bf{24.67} &  18.61 & 23.16 & \bf{9.24}\\\hline
 DW-RNN & 22.52  & 25.90  & 18.75  & 24.44 & 21.44\\
 MTL-RNN & 21.47 & 25.78 & \textbf{17.98} & \bf{22.82} & 17.73\\
 \hline
 \end{tabular}
   \label{table:rul_results}
\vspace{-1em}
\end{table}

\section{Results and Analysis}
\label{sec:results}

\begin{table*}[t!]
\centering
 \caption{ AUCROC and AUCPR scores for the failure prediction using different methods.}
 \begin{tabular}{|c|c|c |c|c|c|c|c|c|c|c|} 
 \hline
&  \multicolumn{8}{c|}{C-MAPSS} & \multicolumn{2}{c|}{BackBlaze} \\\hline
 & \multicolumn{2}{c}{FD01} & \multicolumn{2}{|c}{FD02} & \multicolumn{2}{|c}{FD03} & \multicolumn{2}{|c|}{FD04} & & \multicolumn{1}{c|}{} \\ \hline
 Method  & AUCROC & AUCPR  & AUCROC & AUCPR  & AUCROC & AUCPR & AUCROC & AUCPR & AUCROC & AUCPR \\ 
 \hline
 LR & 0.9904 & 0.6115  & 0.9866 & 0.5222  & 0.9805 & 0.4514  & 0.9707 & 0.3429& 0.7205 & 0.0801\\ 
 SVM  & 0.9940 & 0.6327  & 0.9855 & 0.5001  & 0.9780 & 0.4083  & 0.9706 & 0.3429 & 0.7113 & 0.0916\\
 SVR  & 0.8668 & 0.5979  & 0.9075 & 0.5688 & 0.8860 & 0.5455 & 0.7166 & 0.1987 & 0.5000 & 0.0164 \\
  RF  & \bf{0.9977} & \textbf{0.7884}  & 0.9869 & 0.6212  & 0.9950 & 0.7715  & \bf{0.9880} & \bf{0.3848} & \bf{0.8374} & 0.1864 \\\hline
 FP-RNN  & 0.9913  & 0.6656  & 0.9829 & 0.5328  & 0.9930 & 0.5559 & 0.9727 & 0.3441  & 0.7042  & 0.1721 \\
 RUL-RNN  & 0.7281 & 0.6672  & 0.7763 & 0.5214 & 0.7189 & 0.6874 & 0.6377 & 0.1685 & 0.5878 & 0.1039\\\hline
DW-RNN  & 0.9965 & 0.7560  & 0.9859 & 0.6189 & \textbf{0.9978} & \textbf{0.7889} & 0.9751 & 0.3452 & 0.7217 & 0.1982\\
 MTL-RNN  & 0.9968 & 0.7500  & \bf{0.9872} & \bf{0.6220} & 0.9954 & 0.7701  & 0.9752 & 0.3465 & 0.7374 & \bf{0.2612}\\
 \hline
 \end{tabular}
\vspace{-1.5em}
   \label{table:fp_results}
\end{table*}


In this section, we present our results for the models described in Section~\ref{sec:approach} along with our baselines. The results for C-MAPSS and Backblaze datasets for the RUL and failure prediction tasks are reported in Tables~\ref{table:rul_results} and ~\ref{table:fp_results}, respectively.

\subsection{RUL}
We can observe from the Table~\ref{table:rul_results} that RUL-RNN is the best performing method for the RUL prediction. Our proposed DW-RNN and MTL-RNN methods are competitive with RUL-RNN across the C-MAPSS's all four sub-datasets. Amongst them, DW-RNN and MTL-RNN
perform similar on the C-MAPSS dataset. 
All our methods beat the strong baseline regression methods of SVR. 
These results are significant since the RUL-RNN has a dedicated focus on reducing the RMSE of the RUL prediction along with multiple \emph{labels} --- values from max of RUL to filter window size --- while the multi-task learning network 
has to balance it with training the failure prediction task, which does not have very reliable 
ground truth labels. Despite these constraints, our methods are competitive with the RUL-RNN on C-MAPSS (table~\ref{table:rul_results}).  
DW-RNN's performance is particularly encouraging since it trains on a 
parametric likelihood and not the direct RMSE loss, unlike the other methods. Since DW-RNN is competitive with other methods, we can say that the Weibull assumption is a reasonable assumption for the C-MAPSS which directly translates to a 
proportional hazard as given by Eq.~\ref{eq:whazard}. 
However, as we observe on the Backblaze dataset our methods, particularly, DW-RNN
performs poorly compared to RUL-RNN, since the hard disks usually experience a sudden degradation in the alert window
making the failure prediction task much more challenging as reflected in the failure prediction results in the next section.

\noindent \textbf{Analysis of RUL prediction}: To understand the prediction of the models on different RUL horizons, we bin the RUL values into different classes as shown in Figure~\ref{fig:conf_matrix} for the C-MAPPS FD01 dataset.  We depict the confusion matrices of the baseline RUL-RNN and our MTL-RNN (which is similar to DW-RNN). Clearly, our models work better near the failure time, which is not surprising since they optimize for the failure prediction and RUL both, while the baseline RUL-RNN only optimizes for the RUL. At the other end of the spectrum, when the failure is far, our methods tend to shift the predictions towards higher RUL values as reflected in the last three rows of the confusion matrix, where the predicted values lie the most in the higher RUL label. This result is essential for the applications where the near failure prediction is much more critical. The literature shows that RUL prediction methods perform poorly far from the failure as we can observe in our matrices as well. The algorithmic cycles are spent by the RUL prediction models in trying to reduce the errors far from the failure, as a result of which the near failure prediction is not as accurate as it can be. Our methods overcome this problem by jointly optimizing for the long-term RUL and short-term failure prediction problems.
In summary, our proposed methods work better than RUL-RNN near the failure time, which might be more important than correct predictions far from the failure for a number of use cases.

\subsection{Failure Prediction}
On the failure prediction task, both our proposed methods Deep Weibull and Multi-task RNN outperform all the baselines by a considerable margin except for random forest ensemble method which is competitive with our methods as shown in Table~\ref{table:fp_results}.  By comparing the performance of FP-RNN with 
Deep Weibull and Multi-task RNN, we can see that training the failure prediction task with RUL task helps significantly in improving the performance for the failure prediction task, which is essentially a threshold function on the RUL value. However, as we see that using predicted RUL values from the RUL specific methods like SVR, RUL-RNN works relatively poorly on the failure prediction task. One of the key reasons for that is the focus of the RUL estimation task to get the higher spectrum of RUL right. Previous studies~\cite{zheng2017long} have shown these methods perform relatively poorly when the machine is very far from the failure, \ie for the higher end of ground truth RUL. Hence, the method spends most of its computation in trying to get the prediction at a point far from the time-to-failure right, which is not very useful for the failure prediction.  
Using joint learning approach not only helps the network utilize the non-failure data but also helps the algorithm focus more on the times near failure, which is more critical for the predictive maintenance task.
 \emph{For example, predicting a remaining useful life of 120 days when the real value is 100 days is 
of much less consequence than predicting 25 days when the real RUL is 5 days from a decision-making point of view.}
We observe similar results on the Backblaze dataset. However, with a considerable difference in the AUCPR scores, we can say that our method gives more weight to the failure prediction class compared to the baseline methods. The difference in the AUCROC and AUCPR is drastic on the Backblaze due to comparatively poor performance on the failure prediction task across the algorithms, which makes former a much better indicator of the performance of the algorithms.

\subsection{Consistency in the tasks}
In order to evaluate our initial hypothesis that a joint model of RNN and FP would be helpful in providing consistent results, we use the 
 \textbf{\emph{Spearman's rank correlation}} metric on the predicted values of failure prediction and RUL as shown in Table~\ref{table:spr_corr}. Since, 
there is a complementary relationship between the two tasks a perfect consistency would indicate a correlation of -1. Correlations $\ge 
0.6$ are considered strong, while $\ge 0.8$ are considered very strong. As can be observed, our methods give highly consistent predictions for the two tasks with decent performance on the two tasks on the two tested datasets. Multi-task learning method especially provides very strong correlation scores consistently across the datasets while the Deep 
Weibull exhibits strong to very strong correlations. 

These consistency results are significant since our methods are the only methods
that can do the failure prediction and RUL together. The other way is to use the RUL based prediction method's predicted values for the failure prediction task, which as shown in the previous sections performs very poorly. Hence, our methods can give \textbf{\emph{consistent results}} with a small decrease in performance on the RUL prediction task, but a \textbf{\emph{drastic improvement}} on the failure prediction task compared to using RUL predictions for failure prediction. 

\begin{table}[b!]
	\vspace{-1em}
  \caption{Spearman's rank correlation for predicted failure probabilities and RUL values for each method. All the reported correlations have a $p$-value $<0.001$.}
\centering
 \begin{tabular}{|c|c|c|c |c|c|} 
 \hline
Method & {FD01} & {FD02} & {FD03} & {FD04} & BackBlaze \\ \hline
 \hline
 RUL-RNN & -1.000 & -1.000 & -1.000 & -1.000 & -1.000\\
 DW-RNN & -0.548 & -0.827 & -0.612 & -0.732 & -0.657 \\
MTL-RNN & -0.860 & -0.856 & -0.709 & -0.604 & -0.956\\
 \hline
 \end{tabular}
\vspace{-1.5em}
   \label{table:spr_corr}

\end{table}

\subsection{Deep Weibull vs Multi-Task Learning}
The benefits of the Deep Weibull network are multi-fold: it learns the distribution of the time-to-failure event random variable, without explicitly training for particular metrics and hence is much more generalizable. This can help us get much more information like detecting sudden changes in the device's health by analyzing the hazard rates or the survival functions, without any need for a dedicated procedure that would be required in other
processes. Additionally, we can derive multiple failure predictions from the same trained model with different horizon values, that none of the other methods are capable of.
However, the assumption of a Weibull distribution on the random variable might be too strong and not generalize across domains. As we can see in the Tables~\ref{table:rul_results} and~\ref{table:fp_results} that
while Weibull performs well on the C-MAPSS dataset, its performance on the Backblaze dataset is considerably lower than the Multi-task learning. 
 While we change the Weibull parameters based  on the input window, this assumption might not hold in processes that suffer from the sudden degradation like hard disks show, 
with the proportional assumption of the hazard rate in Weibull  (Eq~\ref{eq:whazard}).
Our Multi-task Learning method overcomes the Weibull assumption by learning the underlying distribution  \emph{``implicitly"} to control the relationship between the RUL and failure prediction tasks.   
Additionally, we found that trying to train the Deep Weibull network is not a trivial task and requires special initialization of the network as described in Section~\ref{sec:settings} and appropriate parameter transformations. 

However, the Multi-Task Learning is not as  \textbf{\emph{expressive}} a model as the Deep Weibull that can be used to get the distributional information like survival functions about the whole degradation process or do failure prediction over new horizon windows. Hence, our methods can be deployed based on the requirements of the
domain.



\section{Conclusion}
\label {sec:conclusion}
In this work, we propose approaches to jointly model two classical problems in predictive maintenance of predicting remaining useful life and failure prediction with an aim to provide consistent results. We propose two techniques to jointly predict RUL and failure prediction --- Deep Weibull RNN and Multi-task learning RNN. While the Deep Weibull learns the tasks implicitly using a survival likelihood fitted with a Weibull distribution, multi-task deep network learns the tasks explicitly.  We show that our methods perform better than most of the baselines that are trained for one of the tasks on both the tasks by exploiting the relationship between the two tasks. In particular, our methods do very well on the harder task of failure prediction with a 
subtle drop on the RUL estimation with the comparable deep learning networks. Our multi-task learning framework is different from the traditional multi-task networks since we leverage the non-failure data to train the network with failure prediction task. Both our methods are able to utilize the non-failure data unlike traditional approaches within the training procedure without any need for creating embeddings for pre-training. Our methods provide consistent predictions for both the tasks providing a solution that can be deployed in practical maintenance tasks.

\balance

\bibliographystyle{IEEEtran}
\bibliography{main}




\end{document}